\documentclass[conference]{IEEEtran}
\usepackage{times}

\usepackage[numbers]{natbib}
\usepackage{multicol}
\usepackage[bookmarks=true]{hyperref}
\usepackage{graphicx}
\usepackage{amsmath,amssymb}
\usepackage{mathtools}

\begin{document}

\title{A Conditional Adversarial Network for Scene Flow Estimation}

\author{Ravi Kumar Thakur and Snehasis Mukherjee}

\maketitle

\begin{abstract}
The problem of Scene flow estimation in depth videos has been attracting attention of researchers of robot vision, due to its potential application in various areas of robotics. The conventional scene flow methods are difficult to use in real-life applications due to their long computational overhead. We propose a conditional adversarial network SceneFlowGAN for scene flow estimation. The proposed SceneFlowGAN uses loss function at two ends: both generator and descriptor ends. The proposed network is the first attempt to estimate scene flow using generative adversarial networks, and is able to estimate both the optical flow and disparity from the input stereo images simultaneously. The proposed method is experimented on a huge RGB-D benchmark sceneflow dataset.

\end{abstract}

\IEEEpeerreviewmaketitle

\section{Introduction}
Scene flow is a three dimensional (3D) motion field representation of points moving in the 3D space. Scene flow gives the complete information about the motion and geometry in a stereo pair of frames in 3D space, of all the visible scene points in the frames. Thus, the estimation of the flow field is an important task in 3D computer vision and robot vision. The work on motion estimation has been done earlier for rigid scenes. However, the problem of scene flow estimation started gaining attention when scene flow was first introduced for dynamic scenes \cite{vedula1999three}. This complete understanding of dynamic scene can be used in many application areas of computer vision including activity recognition, 3D reconstruction, autonomous navigation, free-viewpoint video, motion capture system, augmented reality, and structure from motion. The problem of scene flow estimation can be considered as 3D counterpart of optical flow estimation.

\begin{figure}[t]
  \includegraphics[width=\columnwidth]{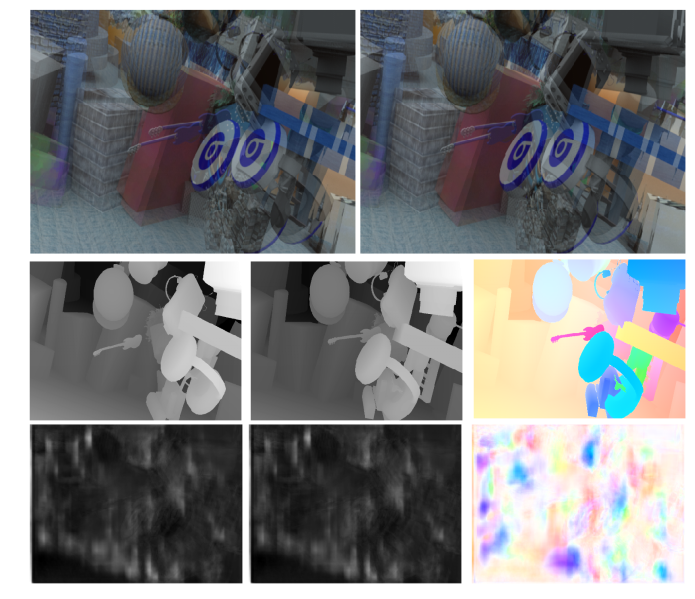}
  \caption{An example of stereo pair alongwith the Ground truth and predicted disparities, optical flow. First row shows a sample stereo pair, second row shows the ground truth scene flow and the third row shows the reconstructed scene flow along x-, y- and z-directions.}
  \label{figure1}
\end{figure}

Despite of several efforts, estimation of scene flow is still an under-determined problem. Several approaches for scene flow estimation have been proposed since its introduction, where most of the approaches rely on conventional procedures of computer vision \cite{huguet2007variational}. Some scene flow estimation methods extend the popular optical flow estimation techniques to 3D by introducing disparity map, for estimating scene flow \cite{yamaguchi2014efficient}. While other approaches are based on optimization of energy function and variational methods \cite{quiroga2014dense}. Most of the scene flow estimation  methods rely on the calculation of optical flow alongwith depth for estimating the scene flow \cite{jaimez2015primal}. The fundamental assumptions behind the state-of-the-art algorithms are brightness and gradient constancy of the stereo images. As a result, most of the methods work very well on scenes with small displacement, but can not perform well on large displacement samples. In realistic scenario, these assumptions are violated often. Figure \ref{figure1} shows such an example from our prediction results.

The problem of scene flow estimation using deep networks has recently attracted the attention of computer vision research community with availability of large scale datasets \cite{mayer2016large}. 
Nearly all the classical methods take several minutes to process a frame. Hence, the computational time does not permit real time application. Recently, learning based methods \cite{DBLP:conf/bmvc/Qiao0LZYX18, mayer2016large} have been proposed due to availability of large scale dataset with ground truth. These methods take more time for training, but are able to reduce the run time to less than a second. Though in terms of accuracy, this learning based approaches can be currently not on par with the classical methods. With the training in synthetic dataset the scene flow estimation may not work on naturalistic scenes.

Estimating scene flow is a challenging problem because of the dependency of the estimation algorithms on the assumptions of brightness and gradient constancy across subsequent stereo frames. Even an occlusion in the image can affect the stereo correspondence between the frames. Varying illumination and lack of texture information can also give erroneous information about the brightness pattern. Similarly, large displacement can also cause error in scene flow computation. The deep networks are good in understanding highly abstract features. The automatic feature extraction capability of the deep networks can be used to develop more robust model for the cases where assumptions are violated. 

There are a few learning based methods for scene flow estimation.  This can be useful in semi-supervised learning scenario as well, since acquiring data will remain a challenge. We propose a conditional adversarial network for estimating scene flow from stereo images obtained at different time instances. To our knowledge this is the first attempt on scene flow estimation using GANs. The presence of additional discriminator network as a critic can direct the training process for scene flow estimation. Thus, the scene flow estimation benefits from this adversarial model. This generative modeling approach can also be used for unsupervised learning of scene flow. At present, there are no large scale dataset with naturalistic scene is available. However, the proposed SceneFlowGAN can be used to generate such datasets. 

\begin{figure*}[t]
  \centering
  \includegraphics[width=\textwidth]{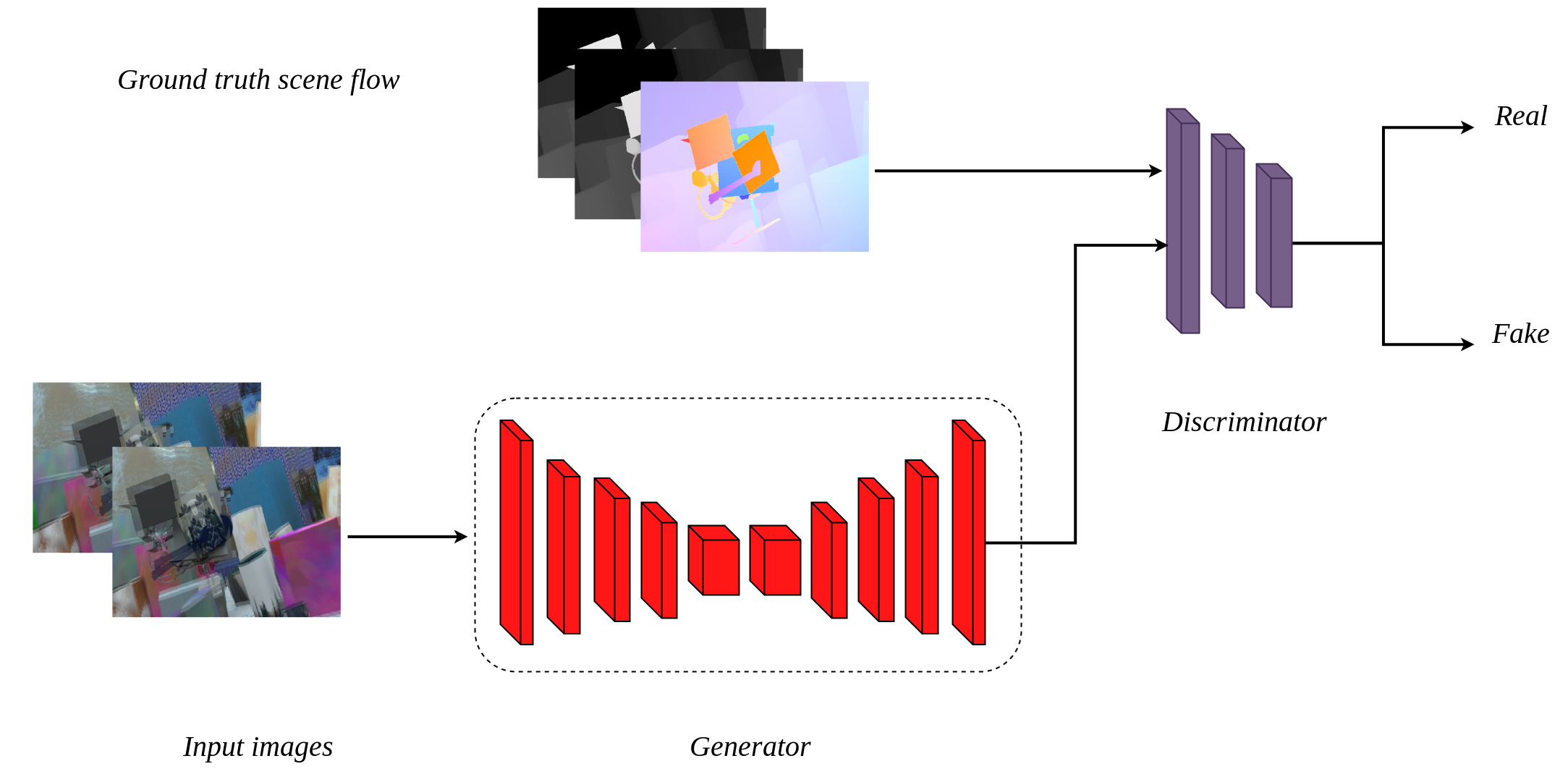}
  \caption{The proposed architecture for SceneFlowGAN. the generator follows an encoder-decoder architecture. The networks are composed of units in the form of convolution-batch norm-leakyReLU. This units are also part of discriminator  network }
  \label{figure2}
\end{figure*}

\section{Related Works and Background}
Scene flow estimation using deep networks is an active area of research \cite{ISKB18}. We discuss recent advances in scene flow estimation, generative adversarial networks and their applications in structure prediction problems in separate subsections. 

\subsection{Scene Flow Estimation}
Classical scene flow estimation methods are generally based on data extracted from sequence of images obtained from multiple cameras, stereo images and depth data. Scene flow  was first proposed by \citet{vedula1999three} using multi-view images. They obtained multi-view scene flow from optical flow and surface geometry. Usually such methods use some 3D reconstruction procedure. Scene flow based on stereo image from binocular setting often involves joint estimation of optical flow and disparity \cite{huguet2007variational, yamaguchi2014efficient}. Though, some scene flow estimation methods are based on stereo images decoupled into stereo and motion estimation \cite{wedel2011stereoscopic}. \citet{basha2013multi} formulated the structure and scene flow in point cloud representation. Scene flow by enforcing depth discontinuity using image segmentation information was introduced by \citet{zhang20013d}. It computes both the 3D motion and the 3D structure. Most of these methods uses variational framework. However, \citet{schuster2018sceneflowfields} proposed scene flow estimation method based on dense interpolation of sparse matches from stereo images. The variational optimization was used at later stage for refinement. 

With the advent of depth cameras, scene flow estimation methods using RGB-D data were explored \cite{quiroga2014dense, herbst2013rgb, jaimez2015primal}. However, the depth cameras are not sufficiently accurate in outdoor environment. They pose limitation due to changes in illumination, frame rate and limited field of view. The classical methods of estimating motion rigidly follow the brightness and gradient constancy assumption. However, most of these assumptions do not hold true in dynamic environments. 

\subsection{Structure and Motion Estimation from Deep Networks}

For motion estimation based on deep networks, the availability of large dataset was a challenge. Since acquiring motion data for naturalistic scene was tedious, \citet{mayer2016large} introduced FlyingThings3D synthetic dataset. Recently, motion estimation based on deep network have shown the promise of such methods. The introduction of FlyingThings3D dataset gave boost to such CNN based methods for motion estimation. They were also the first to apply CNN for scene flow estimation by proposing SceneFlowNet \citet{mayer2016large}. SceneFlowNet used combined architecture of FlowNet\cite{Dosovitskiy_2015_ICCV} and DispNet\cite{mayer2016large} for estimating scene flow. This was subsequently revised in   FlowNet2 \cite{ilg2017flownet}. They addressed the problem of large displacement by stacking different architectures of FlowNet. The small displacement was addressed using small strides in convolution layers. SpyNet \cite{ranjan2017optical} used spatial pyramid of input data to reduce the number of training parameters. 

Motivated by the success in estimating optical flow through CNN, a few deep networks for scene flow estimation were also proposed. \citet{ISKB18} introduced stacked architecture based on FlowNet2.0 to estimate disparity and scene flow in occluded stereo images. \citet{Behl_2017_ICCV} combined recognition with geometry information to estimate scene flow in dynamic scene with large displacement. SF-Net \cite{DBLP:conf/bmvc/Qiao0LZYX18} introduced end-to-end training for scene flow estimation from RGB-D images. A CNN for direct estimation of scene flow was proposed by  \citet{thakur2018sceneednet}. The model SceneEDNet \cite{thakur2018sceneednet} estimates three dimensional motion from temporal sequence of stereo images, without giving geometry information. \citet{vijayanarasimhan2017sfm} solved for 3D motion and 3D geometry simultaneously by using two different networks for structure and motion. The SceneEDNet is a deep network for end-to-end learning of sceneflow using only stereo images, which can be fed into a GAN readily. Hence, we use the SceneEDNet architecture in the Generator part of the proposed SceneFlowGAN architecture.

\subsection{Generative Adversarial Networks}
The work on Generative Adversarial Network was proposed by \citet{goodfellow2014generative}. The GAN architecture consists of two networks training in an adversarial mode against each other. The generator is tasked to generate realistic images given a latent noise sample. While, the discriminator network is supposed to train on both real as well generated image so to be able to distinguish between the two. Both, the generator and the discriminator are involved in a min-max game.  This can be represented by following equation.
\begin{equation}
    \min_{G} \max_{D} \mathbb{E}_{x~P(x)}[log(D(x)] + \mathbb{E}_{z~P(z)}[log(1-D(G(z))] 
\end{equation} 

The generator is denoted by $G$ and discriminator by $D$. $P(z)$ is the distribution of noise Training of GAN has been difficult due to problems such as vanishing gradient, mode collapse. The training can also be highly unstable. Wasserstien GAN \cite{arjovsky2017wasserstein} overcame some of these challenges by using EM or earth mover's distance as loss function. Also, in some cases the discriminator and generator loss values are not good indicator of training of GAN. \citet{mirza2014conditional} proposed conditioning of both the generator and the discriminator on additional information available with the data. This allowed to direct the training of GAN for data generation. 
These advances in GAN were followed by its application in various area of computer vision. \citet{kupyn2018deblurgan} demonstrated a conditional adversarial network for restoring a blurred image. They used residual blocks for generator with perceptual loss. \citet{Zhang_2017_ICCV} synthesized images from text description from stack of two GANs. The first GAN generates rough images based on text description. The second GAN is conditioned on first one to perform refinement. Generation of super-resolution from single image was achieved by \citet{ledig2017photo} using perceptual loss. A model for image to image translation \cite{isola2017image} was proposed by conditioning both the adversarial networks on input image. A semi-supervised optical flow estimation using conditional GAN was proposed \cite{lai2017semi} using both labeled and unlabeled data.

\section{Proposed Method}

We propose a conditional adversarial network for estimating scene flow from pairs of stereo images. The weights of the generator and discriminator are updated together during the training phase. The learning of optical flow and disparity are coupled in SceneFlowGAN.

\subsection{Scene Flow Estimation}
Given stereo image pairs at consecutive time instances the scene flow can be constructed from optical flow ($u, v$) and disparity ($d_{t}, d_{t+1}$). The dense scene flow provides 3D position and the constituent 3D motion vector for all the points. Our proposed method takes set of stereo images defined by $\mathcal{I}=(I_{L}^{t}, I_{L}^{t+1}, I_{R}^{t}, I_{R}^{t+1})$ to generate scene flow $\mathcal{S}$. Thus, the scene flow can be considered as a 4D vector.
\[ \mathcal{S} = (u, v, d_{t}, d_{t+1})\].
The horizontal and vertical components of optical flow is represented by $u$ and $v$ respectively. Disparities of stereo pairs at $t$ and $t+1$ are denoted by $d_{t}$ and $d_{t+1}$.  In point cloud the scene flow can be computed using the camera parameters and pinhole projection model. When projected on the image plane, the scene flow gives corresponding optical flow. 

\subsection{Adversarial Training}
For training SceneFlowGAN, the loss function is computed twice, one at the end of discriminator and other at the generator's. The discriminator's loss makes the network learn to identify ground truth and generated scene flow. The discriminator is not conditioned like the one proposed in \cite{mirza2014conditional}. The loss at the end of generator $\mathcal{G}$ makes sure that network is optimized for scene flow estimation task from pair of stereo images. At the same time, the generator is also trained to pass the critic test by discriminator. This one-to-many mapping directs the training of generator for scene flow estimation task.

\[ \mathcal{L} = \mathcal{L}_{GAN} + \mathcal{L}_{Joint Loss}\]

For training the generator we define a joint loss function. It is composed of average end point error for optical flow and an $L1$ loss for calculating the error between the two disparity values. The optical flow error is the average end-point-error. The error in disparity is given by L1 loss. We use wasserstein metric as GAN loss function for stable training using gradient descent\cite{arjovsky2017wasserstein} . The joint loss function can be given as below

\begin{equation}
\begin{aligned}
    \mathcal{L}_{joint loss} = & \Sigma\sqrt{(u-u')^2 + (v-v')^2} + \\
    & \Sigma\left(|{d_{t}- d'_{t}}|\right) +
    \Sigma\left(|d_{t+1}- d'_{t+1}| \right) 
\end{aligned}
\end{equation} 

The GAN loss takes the decision on input scene flow being real or generated. The conditioning of generator on input stereo images makes generator learn to estimate scene flow and also to fool the discriminator.  Thus, the SceneFlowGAN is trained to optimize the following objective function 
 
 \begin{equation}
     \min_{G} \max_{D} \mathcal{L}_{GAN}(G,D) + \mathcal{L}_{Joint Loss}(G)  
 \end{equation}

The discriminator network tries to maximize the objective function while the generator tries to minimize it.

\subsection{Architecture of the Proposed Model}
The architecture of SceneFlowGAN is shown in \ref{figure2}. The model consists of a generator and discriminator network. Both the networks are convolutional. For the generator we have used SceneEDNet\cite{thakur2018sceneednet}. Unlike \cite{thakur2018sceneednet} we have used batch-normalization layers for regularization. The network needs to learn correspondences between the stereo pairs. However, much information is lost while propagating from encoding to decoding stage. Thus, we have used skip connections between layers with same dimension in the encoder and decoder part. 

The composition unit of the generator and discriminator networks are of the form convolution-batch normalization-leakyReLU. The discriminator network has three convolution layer each followed by batch-normalization and leakyReLU. The final convolution layer is flattened to connect to set of three dense layers. There is dropout layer with value of 0.4. The last dense layer gives probability value of the scene flow being generated or real.

\section{Experiments and Results}
We describe the datasets being used for training of SceneFlowGAN followed by implementation and results.

\subsection{Dataset}

For training SceneFlowGAN we have used the large scene flow dataset \citet{mayer2016large}. The dataset is divided into three sections, FlyingThings3D, Monkaa and Driving. All the datasets provides 3D scene points. The 3D models were used to create frame artificially used Blender. The scenes are rendered in a way to provide variation in orientation and position for all the visible scene points. The datasets comes with bi-directional optical flow and bi-directional disparity ground truths. The stereo images are available in two formats. One is clean pass, with no noise or external effects. Other is final pass, which comes with motion blur, illumination effects and image degradations. For training we have used FlyingThings3D dataset with final pass images.  

\subsection{Implementation Details}

The estimated scene flow is conditioned on the input stereo pairs at consecutive time instances. For the generator $G$ architecture we have used SceneEDNet\cite{thakur2018sceneednet} with skip-connection. The discriminator $D$ is unconditoned, which is trained to distinguish between generated and ground truth scene flow. During the training, both the network are trained in adversarial manner.

\begin{figure}
  \centering
  \includegraphics[width=8cm]{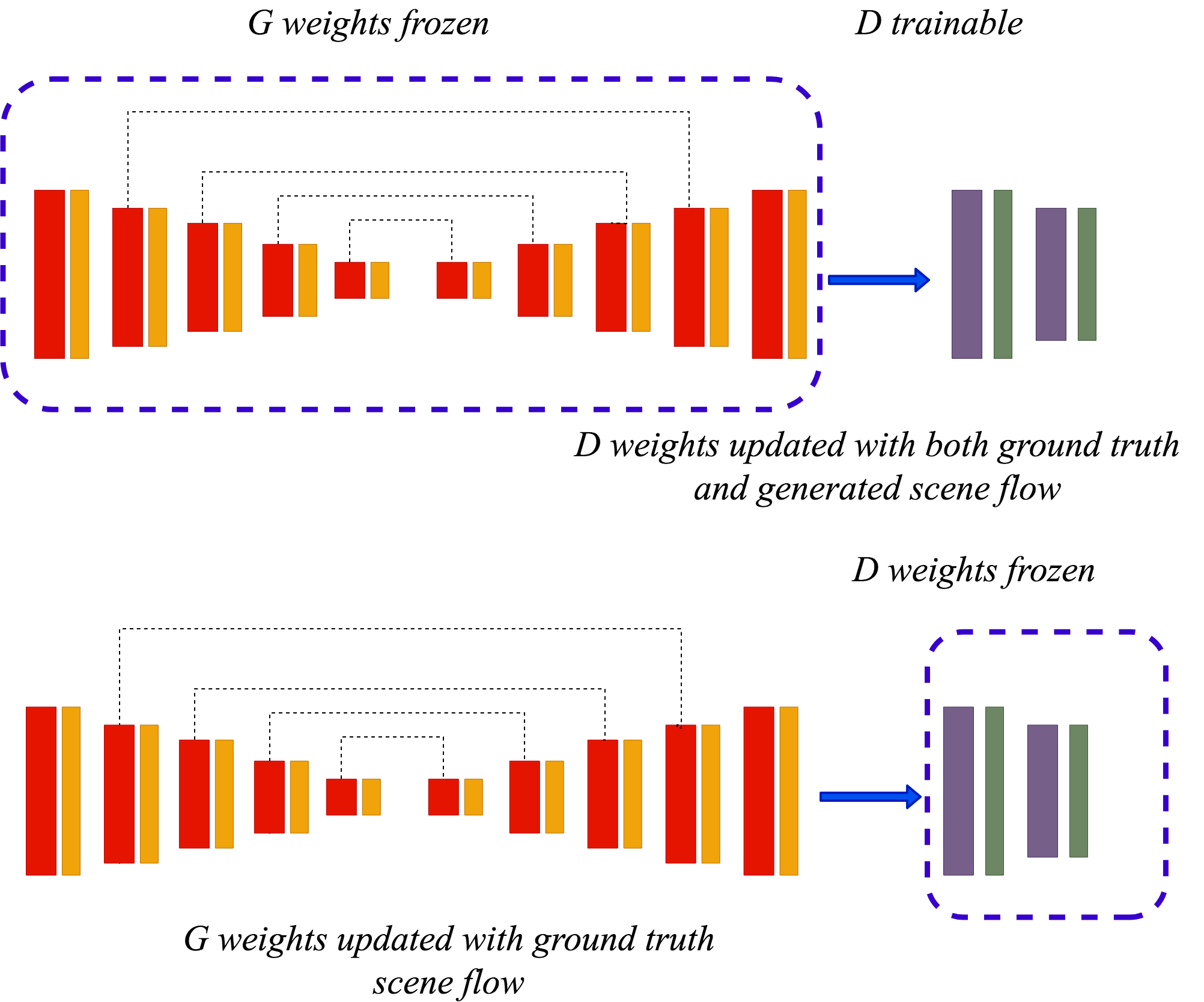}  \caption{Training procedure for SceneFlowGAN. The discriminator and generator are trained in alternating manner.}
  \label{figure3}
\end{figure}

For training the SceneFlowGAN we follow the procedure mentioned in original work\cite{goodfellow2014generative} as shown in \ref{figure3}. We alternate between the training of discriminator and the generator. The discriminator network is trained on both, the ground truth and the generated scene flow. The generator is trained via GAN by making the weights of discriminator frozen. Both the network were trained with Adam\cite{kingma2014adam} optimizer with learning rate of 1e-5. The calculation of loss happens at two places, one at each of generator and discriminator's end. All the training were performed on NVIDIA-1080 GPUs.

\subsection{Results}

\begin{figure*}[t]
  \centering
  \includegraphics[width=\textwidth]{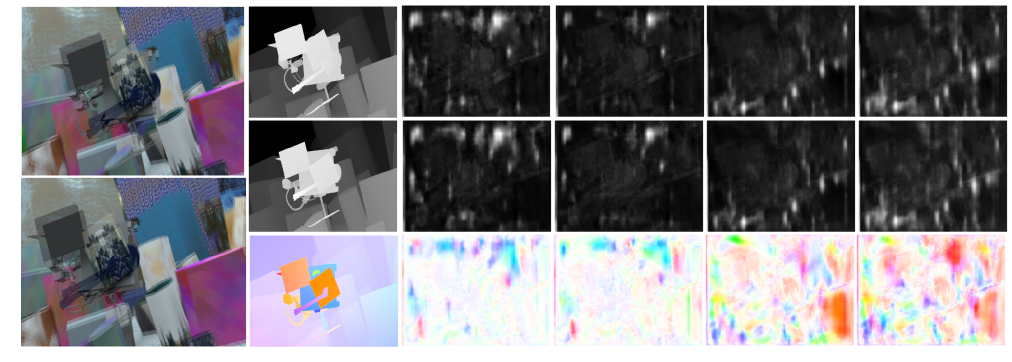}
  \caption{The predicted scene flow from SceneFlowGAN trained on set A and C of FlyingThinsg3D for a pair of stereo images. From left to right. Left and right stereo pair overlaid, ground truth disparity and optical flow. The predictions (from left to right) are for
  SceneFlowGAN trained on set A for 70 epcohs and 50 epcohs, trained on set C for 70 and 50 epochs respectively. }
  \label{figure4}
\end{figure*}

The SceneFlowGAN was trained on FlyingThings3D\cite{mayer2016large} dataset. The learning of optical flow and disparity are coupled. For a input pairs of $\mathcal{I}=(I_{L}^{t}, I_{L}^{t+1}, I_{R}^{t}, I_{R}^{t+1})$ we have obtained corresponding optical flow and disparities for consecutive time instances. The model was trained on final pass stereo images with added image degradations. The results on a stereo pair is showed in \ref{figure4}

\subsection{Ablation Studies}
\begin{figure}
  \centering
  \includegraphics[width=8cm]{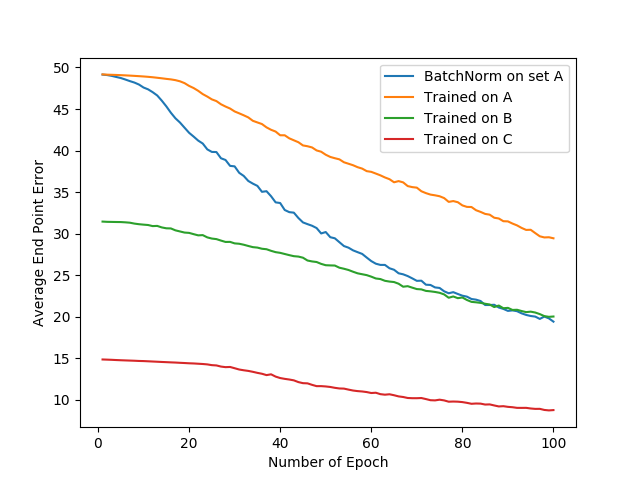}
  \caption{Training loss of SceneEDNet on three sets of FlyingThings3D scene flow. The original SceneEDNet does not have batch-normalization layers. We found decrease in training loss its introduction.}
  \label{figure5}
\end{figure}

The choice of dataset for training was based on training performance of the SceneEDNet \cite{thakur2018sceneednet} on three sets of FlyingThings3D. The training loss curve for SceneEDNet is given in \ref{figure5}. The drop in the average end point error was more for set-B and set-C as compared to A. Moreover, we also observed the drop in the loss value due to additional batch-normalization layers. We trained our model on set-A and set-C of FlyingThings3D scene flow data.  This was done to see the effect of data distribution on learning the generator. \ref{tab:Experiments} shows the flow and disparity error obtained by SceneFlowGAN on all the test sets of FlyingThings3D. The test was done for both the models trained on A and C.

\begin{table*}[t]
\centering
\centering 
\resizebox{\textwidth}{!}{
\begin{tabular}{|l|l|l|l|l|l|l|l|l|l|l|l|l|}

\hline
  & \multicolumn{3}{l|}{SceneFlowGAN-A(70)} & \multicolumn{3}{l|}{SceneFlowGAN-C(70)} & \multicolumn{3}{l|}{SceneFloGAN-A(50)} & \multicolumn{3}{l|}{SceneFlowGAN-C(50)} \\ \hline
  & Flow      & d\_1      & d\_2      & Flow      & d\_1      & d\_2      & Flow      & d\_1      & d\_2      & Flow      & d\_1      & d\_2      \\ \hline
A & 72.33     & 33.68     & 32.82     & 72.11     & 37.37     & 39.12     & 71.50     & 35.61     & 35.33     & 72.27     & 36.29     & 37.89     \\ \hline
B & 33.89     & 31.15     & 29.73     & 28.99     & 34.09     & 34.91     & 31.13     & 32.67     & 32.07     & 29.18     & 33.16     & 34.12     \\ \hline
C & 25.18     & 32.72     & 30.89     & 19.06     & 35.40     & 35.66     & 22.08     & 34.28     & 32.68     & 19.86     & 34.54     & 34.86     \\ \hline

\end{tabular}}
\caption{Flow and disparity error obtained for SceneFlowGAN. The error values are obtained after testing both the trained models (A, C) on test set (A, B, C). The value in the bracket afterr model name shows number of epochs.}
    \vspace{5mm} 
\label{tab:Experiments}
\end{table*}

\section{Conclusions} 
In this paper we have a presented a conditional generative adversarial network to estimate the scene flow from stereo images. The training of the SceneFlowGAN remains a challenge given the complexity of the problem. The choice of generator was dependent on the training loss obtained in training the generator separately. In future the proposed GAN based scene flow estimation method can be extended for applying on naturalistic images after creating a sufficiently large dataset, which may lead to a new direction of research on flow field estimation.

\bibliographystyle{plainnat}
\bibliography{references}

\end{document}